\documentclass[runningheads]{llncs}
\usepackage[T1]{fontenc}
\usepackage{graphicx}
\usepackage{orcidlink}
\usepackage{amsmath}
\usepackage{amssymb}
\usepackage{subfigure}
\usepackage{booktabs}
\usepackage{float}
\usepackage{hyperref}
\usepackage{eurosym}
\begin{document}
\title{EyeTheia: A Lightweight and Accessible Eye-Tracking Toolbox\thanks{The French State under the France-2030 programme and the Initiative of
Excellence of the University of Lille are acknowledged for the funding and
support granted to the R-CDP-24-005-CALYPSO project. This work was partially supported by the “CAPES-COFECUB” programme (project number: PE3335662P), funded by the French Ministry for Europe and Foreign Affairs, the French Ministry for Higher Education and CAPES. Research conducted within the context of the {\em International Associated Laboratory}  SAVOIR 21 funded by the University of Lille.}}

%
%
    \author{
    Stevenson Pather\inst{1}\orcidlink{0009-0000-3405-729X} \and
    Niels Martignène\inst{1}\and
    Arnaud Bugnet\inst{1}\orcidlink{0009-0001-6634-897X} \and
    Fouad Boutaleb\inst{1,2}\orcidlink{0009-0009-1253-6295} \and
    Fabien D'Hondt\inst{1,3}\orcidlink{0000-0001-5683-0490} \and
    Deise Santana Maia\inst{2}\orcidlink{0000-0002-8886-0093}
    }
    
    \authorrunning{Pather et al.}
    
    \institute{
    Univ. Lille, Inserm, CHU Lille, U1172 - LilNCog - Lille Neuroscience \& Cognition, F-59000 Lille, France
    \and
    Univ. Lille, CNRS, Centrale Lille, UMR 9189 CRIStAL, F-59000 Lille, France 
    \and
    Centre national de ressources et de résilience (CN2R), F-59000 Lille, France\\
    \email{stevenson.pather@univ-lille.fr,  
    niels.martignene@protonmail.com,
    arnaud.bugnet@chu-lille.fr,
    fouad.boutaleb@univ-lille.fr,
    fabien.d-hondt@univ-lille.fr,
    deise.santanamaia@univ-lille.fr
    }
    }

\maketitle              
\begin{abstract}
We introduce \emph{EyeTheia}, a lightweight and open deep learning pipeline for webcam-based gaze estimation, designed for browser-based experimental platforms and real-world cognitive and clinical research.
EyeTheia enables real-time gaze tracking using only a standard laptop webcam, combining MediaPipe-based landmark extraction with a convolutional neural network inspired by iTracker and optional user-specific fine-tuning.
We investigate two complementary strategies: adapting a model pretrained on mobile data and training the same architecture from scratch on a desktop-oriented dataset.
Validation results on MPIIFaceGaze show comparable performance between both approaches prior to calibration, while lightweight user-specific fine-tuning consistently reduces gaze prediction error.
We further evaluate EyeTheia in a realistic Dot-Probe task and compare it to the commercial webcam-based tracker SeeSo SDK.
Results indicate strong agreement in left--right gaze allocation during stimulus presentation, despite higher temporal variability.
Overall, EyeTheia provides a transparent and extensible solution for low-cost gaze tracking, suitable for scalable and reproducible experimental and clinical studies.
The code, trained models, and experimental materials are publicly available.\footnote{
\href{https://github.com/patherstevenson/EyeTheia}{EyeTheia repository} and
\href{https://git.interactions-team.fr/INTERACTIONS/calypso/src/branch/main/src/lonely_tester}{Calypso experimental platform}.}

\end{abstract}

\section{Introduction}

Eye tracking is a widely used experimental technique in cognitive neuroscience, psychology, and human–computer interaction, providing fine-grained measurements of visual attention and underlying cognitive processes~\cite{rayner1998eye}.
However, laboratory-grade eye-tracking systems typically rely on dedicated hardware, controlled recording environments, and specialized expertise, which limit scalability and complicate deployment in real-world clinical or remote research settings.

Recent advances in computer vision and deep learning have enabled gaze estimation using standard RGB cameras, opening new perspectives for low-cost and accessible eye tracking.
However, many existing solutions remain difficult to integrate into experimental workflows, rely on proprietary components, or lack transparency regarding training data and internal processing.
In parallel, the rise of browser-based experimental platforms places strong constraints on real-time performance, adaptability to individual users, and robustness under unconstrained recording conditions.
Importantly, the requirements of many cognitive paradigms emphasize relative gaze allocation and temporal dynamics over sub-degree spatial accuracy.
Yet, most webcam-based gaze estimation benchmarks continue to focus on absolute spatial precision, leaving open the question of their suitability for demanding experimental paradigms with tight temporal constraints.

In this work, we introduce \emph{EyeTheia}, an open and lightweight gaze estimation toolbox designed to address these requirements.
EyeTheia combines browser-based capture and landmark detection with a deep learning backend inspired by iTracker, enabling real-time gaze estimation from a standard laptop webcam.
The framework supports user-specific calibration through lightweight fine-tuning and can be deployed locally within experimental platforms without specialized hardware.

We evaluate EyeTheia through (i) training and calibration analyzes on MPIIFaceGaze, (ii) deployment in a demanding experimental task based on the Dot-Probe paradigm, and (iii) comparison with a commercial webcam-based eye tracker.
Our results show that EyeTheia reliably captures the coarse attentional signals required by established cognitive paradigms while remaining transparent, adaptable, and suitable for low-cost deployment.

\section{Related Work}
\label{sec:related_work}

\subsection{Appearance-Based Gaze Estimation}

Appearance-based gaze estimation commonly relies on convolutional neural networks that regress gaze direction from eye or face images.
A representative approach is \emph{iTracker}~\cite{krafka2016eye}, which combines eye crops, a full-face image, and a binary face grid to predict gaze in a camera-centered metric space.
Trained on the large-scale GazeCapture dataset, iTracker remains a widely used baseline for gaze estimation in unconstrained environments and is particularly suited to calibration-based adaptation.

For laptop-based settings, MPIIGaze~\cite{zhang2015appearance} introduced appearance-based gaze estimation from webcam images.
Lightweight architectures such as AGE-Net~\cite{agenet} achieve competitive angular accuracy while remaining computationally efficient.
However, AGE-Net predicts gaze angles rather than screen-space coordinates, which limits its direct applicability to screen-based cognitive paradigms without additional geometric calibration.

Recent work has explored deeper architectures based on residual networks and Transformer models to improve gaze estimation accuracy, especially for 3D gaze under wide pose variations.\cite{gaze_transformer}
While these approaches achieve state-of-the-art performance, they typically require substantial computational resources and GPU acceleration, making them unsuitable for real-time, browser-based deployment on consumer laptops.
In contrast, our work prioritizes architectural simplicity and deployability over marginal gains in absolute accuracy.

\subsection{Datasets for Gaze Estimation on Consumer Devices}

The effectiveness of appearance-based gaze estimation strongly depends on the alignment between training data and the target application, including camera placement, viewing distance, and screen geometry.
Datasets collected under mismatched conditions often generalize poorly to new usage scenarios.

GazeCapture~\cite{krafka2016eye} provides large-scale, diverse data collected on mobile devices, but its capture geometry differs substantially from that of desktop and laptop usage.
In contrast, MPIIGaze~\cite{zhang2015appearance} and its extension MPIIFaceGaze~\cite{zhang2017s} were explicitly collected using laptop webcams in everyday environments.
MPIIFaceGaze provides full-face images and gaze annotations expressed directly in screen coordinates, making it well aligned with screen-space gaze estimation on consumer laptops.
Its subject-wise structure further enables rigorous subject-disjoint evaluation.

\begin{table}[t]
\centering
\renewcommand{\arraystretch}{1.2}
\resizebox{\linewidth}{!}{
\begin{tabular}{lccccc}
\toprule
\textbf{Dataset} & \textbf{Task} & \textbf{Camera} & \textbf{Free head} & \textbf{Associated model} & \textbf{Model size} \\
\midrule
MPIIGaze        & 2D screen (angles) & Webcam & Yes & AGE-Net~\cite{agenet} & $\sim$4M \\
GazeCapture     & 2D screen (cm)     & Smartphone & Yes & iTracker~\cite{krafka2016eye} & $\sim$95M \\
ETH-XGaze       & 3D / wide pose     & HD camera & Yes & ETH-XGaze~\cite{zhang2020eth} & $>$20M \\
Gaze360         & 3D free            & RGB camera & Yes & MCGaze~\cite{gaze360} & $\sim$18M \\
EYEDIAP         & 3D + depth         & RGB-D & No & RecurrentGaze~\cite{eyediap} & $>$25M \\
RT-GENE         & 3D / real-time     & Multi-camera & No & RT-GENE Ensemble~\cite{rtgene} & $>$100M \\
\bottomrule
\end{tabular}
}
\caption{Comparison of commonly used gaze estimation datasets and their associated model complexity.}
\label{tab:datasets_comparison}
\end{table}

As summarized in Table~\ref{tab:datasets_comparison}, MPIIFaceGaze offers a favorable trade-off between task relevance, data realism, and computational efficiency for screen-space gaze estimation on consumer laptops; therefore, it is used as the primary dataset in this work.

\subsection{Web-Based and Low-Cost Eye Tracking}

Several systems aim to enable eye tracking using standard webcams and browser technologies.
Web-based approaches, such as WebGazer.js~\cite{papoutsaki2016webgazer}, rely on facial landmarks but typically exhibit limited accuracy and temporal stability.
Commercial SDKs, such as SeeSo~\cite{SeeSoSDK}, achieve higher performance through proprietary models at the cost of limited transparency and reproducibility.

To address all the limitations discussed in the related work section, this work contributes:
(i) an open, hybrid gaze estimation pipeline combining browser-based capture with a documented CNN backend;
(ii) an analysis of complementary training strategies and user-specific calibration;
and (iii) validation in a demanding cognitive paradigm with a direct comparison to a commercial tracker.
This design targets low-cost, local deployment in cognitive and behavioral research while maintaining transparency and reproducibility.

\section{Methodology}
\label{sec:methodology}

EyeTheia relies on a complete processing pipeline spanning data preprocessing, feature extraction, model training, real-time inference, and user-specific calibration. The system is built upon the iTracker architecture and investigates two complementary strategies: (i) adapting a model pretrained on GazeCapture, and (ii) training the same architecture from scratch on MPIIFaceGaze to obtain a desktop-oriented baseline.

\subsection{Overview of the Calibration Pipeline}

Before detailing the two learning strategies, we summarize the calibration process common to all variants of EyeTheia. Each calibration session consists of presenting 13 screen targets to the user; for each fixation, a single RGB frame is captured and processed through the feature-extraction pipeline (eyes, face, face grid). The corresponding screen coordinate provides the supervision signal. These pairs of \{inputs, target\} form a compact user-specific dataset used to fine-tune the model.

\begin{figure}[t]
\centering
\includegraphics[width=0.7\linewidth]{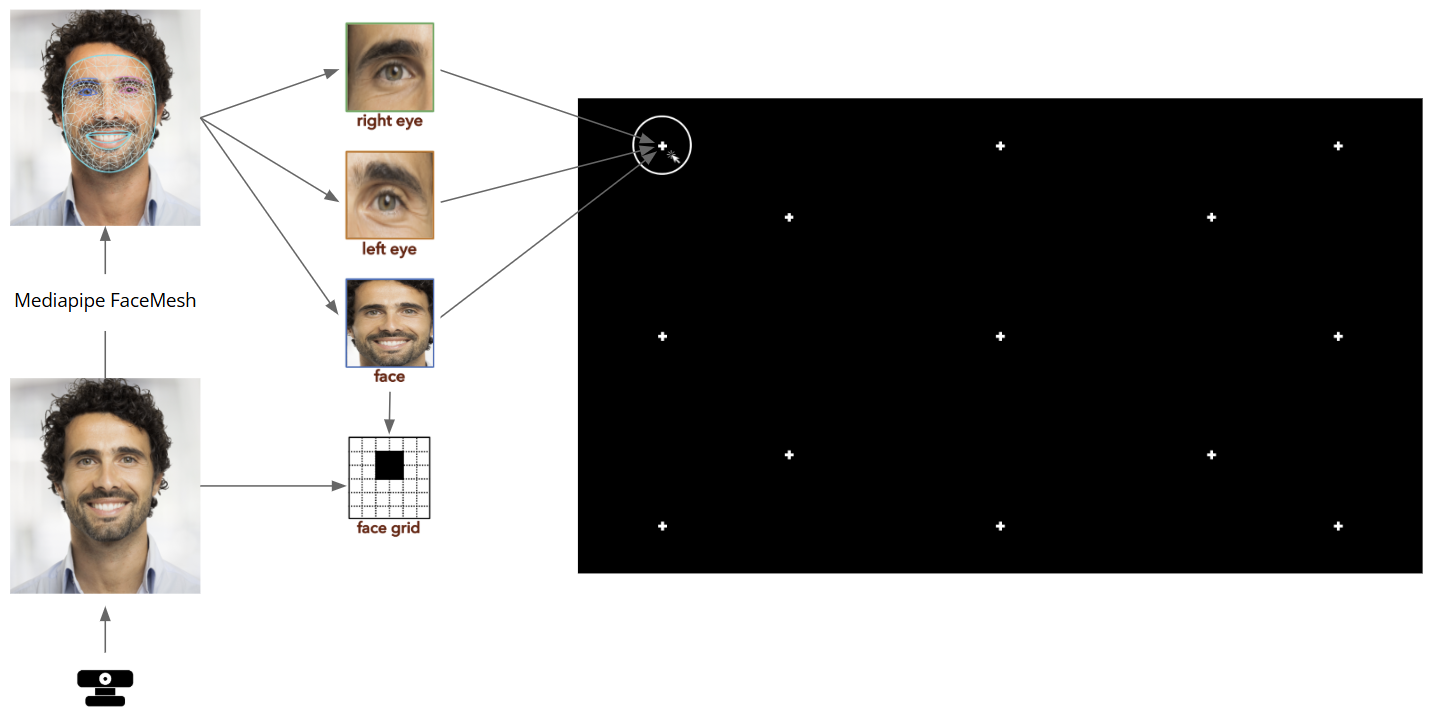}
\caption{Overview of the user calibration pipeline used in both approaches. MediaPipe FaceMesh extracts facial landmarks, from which the iTracker inputs (eyes, face, face grid) are derived. The collected samples are used to fine-tune the model for the current user.}
\label{fig:calibration_pipeline}
\end{figure}

\subsection{Architecture}

We adopt the iTracker architecture introduced by Krafka et al.~\cite{krafka2016eye}, which combines three input streams: (1) left and right eye crops, (2) a full-face crop, and (3) a binary face grid encoding the location of the face in the image. Each stream is processed by a dedicated branch (shared CNN weights for both eyes, an independent CNN for the face, and a fully connected module for the face grid). Their embeddings are concatenated and passed through two fully connected layers to regress the gaze coordinates.

Eye crops are encoded using four convolutional layers (CONV-E1--E4) followed by a 128-dimensional fully connected layer. The face branch uses five convolutional layers (CONV-F1--F5) followed by two fully connected layers (128 and 64 units). The face grid is processed through two fully connected layers (256 and 128 units). The fusion module maps the concatenated features to a 2D output.

\paragraph{Design choices.}
The use of iTracker, MediaPipe FaceMesh, and a 13-point calibration protocol is motivated by practical constraints and alignment with established eye-tracking practices.
iTracker provides a lightweight multi-branch architecture suited for gaze estimation and user-specific calibration.
MediaPipe FaceMesh enables robust landmark-based extraction of facial regions of interest in unconstrained conditions while remaining compatible with real-time browser deployment.
The 13-point calibration follows standard protocols used in commercial systems, balancing spatial coverage and user burden.

\subsection{Feature Extraction via MediaPipe}

In the original iTracker pipeline~\cite{krafka2016eye}, face and eye crops are obtained using a standard Dlib-based face detector, from which bounding boxes are resized to fixed resolutions. A binary face grid encodes the coarse location of the face.

In EyeTheia, this extraction procedure is replaced by a landmark-driven approach using MediaPipe FaceMesh~\cite{lugaresi2019mediapipeframeworkbuildingperception}. For each RGB frame, the predicted landmarks define tight and stable bounding boxes for the eyes and the full face. Each region of interest is resized to the iTracker resolutions (112$\times$112 for eyes, 224$\times$224 for the face). Pixel intensities are normalized to $[0,1]$, and branch-specific mean images (face, left eye, right eye) stored in \texttt{.mat} files are subtracted to ensure compatibility with the original architecture.

The face grid is generated following the iTracker specification by projecting the face bounding box into a $25\times25$ binary mask that encodes its location within the frame.

\subsection{Real-Time Inference Pipeline}

The same feature-extraction pipeline used during calibration is applied to every frame during real-time inference. For each incoming RGB image, MediaPipe FaceMesh extracts facial landmarks from which the three iTracker inputs (eyes, face, face grid) are computed and forwarded to the selected model, producing gaze predictions at the webcam frame rate.

To support browser-based experiments and external applications, EyeTheia exposes this inference pipeline through a lightweight FastAPI server. Two REST endpoints are provided: (i) a calibration endpoint that receives the user’s mini-dataset and triggers fine-tuning, and (ii) a prediction endpoint that accepts individual frames and returns gaze estimates. This design preserves the exact preprocessing and inference steps used in offline experiments while enabling seamless deployment in web and clinical environments.

\subsection{Approach 1: Pretrained iTracker}

The first strategy initializes the network with iTracker weights pretrained on the GazeCapture dataset. Because GazeCapture was collected on mobile devices, predictions are made in a camera-centered metric space:
\begin{equation}
(\hat{x}_{\mathrm{cm}}, \hat{y}_{\mathrm{cm}}) \in [-25, 25]^2.
\end{equation}

\paragraph{Calibration protocol.}
At the beginning of each session, the user fixes 13 screen targets. For each fixation, an RGB frame is processed and its corresponding target is mapped to the metric prediction space.

\paragraph{Fine-tuning.}
All network parameters are updated using Adam ($10^{-4}$ learning rate) and a Euclidean loss:
\begin{equation}
\mathcal{L}(y,\hat{y}) =
\frac{1}{N}\sum_{i=1}^{N}
\big\lVert
\mathbf{y}_i - \hat{\mathbf{y}}_i
\big\rVert_2^2,
\qquad
\mathbf{y}_i =
\begin{bmatrix}
x^{(i)}_{\mathrm{cm}} \\[2pt]
y^{(i)}_{\mathrm{cm}}
\end{bmatrix},
\quad
\hat{\mathbf{y}}_i =
\begin{bmatrix}
\hat{x}^{(i)}_{\mathrm{cm}} \\[2pt]
\hat{y}^{(i)}_{\mathrm{cm}}
\end{bmatrix}.
\end{equation}

\paragraph{Conversion to screen pixels.}
Predictions expressed in centimeters are converted to screen-space pixel coordinates.
Let $W$ and $H$ denote the screen width and height in pixels, respectively.
Following the original iTracker formulation, the conversion is given by:
\begin{equation}
x_{\mathrm{px}} = \frac{25 + x_{\mathrm{cm}}}{50}\, W,
\qquad
y_{\mathrm{px}} = \frac{25 - y_{\mathrm{cm}}}{50}\, H.
\end{equation}

\subsection{Approach 2: Training iTracker from Scratch on MPIIFaceGaze}

The second strategy trains the same architecture directly on MPIIFaceGaze, which provides full-face images and screen-space gaze coordinates captured using laptop webcams.

\paragraph{Normalization of gaze targets.}
Because participants use different screen resolutions, gaze coordinates are normalized per subject during training:
\begin{equation}
(x',\,y') = \left( \frac{x}{W_s},\, \frac{y}{H_s} \right) \in [0,1]^2,
\end{equation}
where $(W_s, H_s)$ denotes the screen resolution associated with subject $s$ in the MPIIFaceGaze dataset, which may differ across participants.
During inference, model predictions $(\hat{x}',\hat{y}')$ are mapped back to pixel coordinates through the inverse transformation:
\begin{equation}
(\hat{x},\,\hat{y}) = (\hat{x}' W_s,\, \hat{y}' H_s).
\end{equation}
This ensures that the network outputs are expressed directly in screen-space units, consistent with the experimental protocol.

\paragraph{Subject-disjoint splitting.}
We use a 5-fold \textit{GroupKFold} to ensure strict subject separation between training and validation.

\paragraph{Training setup.}
The model is trained for 15 epochs using Adam, a batch size of 8, and the same per-branch mean subtraction as in Approach~1.

\paragraph{Loss function and hyperparameter tuning.}
Training uses a Smooth L1 (Huber) loss with transition parameter $\beta$:
\begin{equation}
\mathcal{L}(r)=
\begin{cases}
\frac{1}{2\beta}\, r^2 & \text{if } |r| < \beta,\\[4pt]
|r| - \frac{\beta}{2} & \text{otherwise},
\end{cases}
\qquad r = y' - \hat{y}'.
\end{equation}

We perform a grid search over $\beta \in [0.01, 1.0]$ using a fixed learning rate ($10^{-4}$), recording the validation loss across 15~epochs and selecting the best-performing value. With $\beta$ fixed, several learning rates (e.g.\ $10^{-3}$ to $10^{-5}$) are then evaluated. The configuration yielding the lowest validation loss is retained for all subsequent experiments.

\section{Experiments}
\label{sec:experiments}

We assess EyeTheia within a realistic experimental task that has been extensively used in experimental psychology and is commonly coupled with eye tracking to investigate attentional biases in laboratory settings (e.g. ~\cite{refSciRep2020}), including in studies involving clinical populations (e.g.~\cite{refPsycholMed2023}). Rather than proposing a new behavioral paradigm, we examine whether webcam-based eye tracking can reliably support an established and demanding task traditionally coupled with laboratory-grade eye trackers. In particular, EyeTheia is compared against a commercial webcam-based solution (SeeSo), which is currently used in the reference protocol.

\subsection{Experimental Task: Dot-Probe Paradigm}

The experiment relies on a variant of the Dot-Probe Task (DPT)~\cite{macleod1986attentional}, a widely used paradigm to assess attentional bias toward emotionally salient stimuli. In this task, participants are exposed to pairs of stimuli consisting of one emotionally negative and one neutral item, presented symmetrically on either side of a central fixation.

While attentional bias was historically inferred from reaction times, this approach suffers from limited reliability and provides only indirect access to attentional dynamics. In particular, standard response-time–based indices of attentional bias have been shown to exhibit poor psychometric properties, including low internal consistency and test–retest reliability~\cite{schmukle2005,waechter2014,mcnally2019}. 

For this reason, eye tracking has been increasingly coupled with the Dot-Probe Task to directly monitor gaze allocation during stimulus presentation, allowing a fine-grained and temporally resolved characterization of attentional deployment~\cite{armstrong2012,lazarov2019,refPsycholMed2023}.

\subsection{Stimuli}

Affective visual stimuli were used to probe attentional allocation during the task.

\paragraph{Affective scenes (IAPS).}
A subset of images was selected from the International Affective Picture System (IAPS), comprising negative and neutral stimuli matched for low-level visual properties. Negative images depict scenes related to threat, injury, or distress, while neutral images include inanimate objects or everyday situations. The selection procedure and stimulus material follow protocols previously described and validated in eye-tracking studies of attentional bias in healthy individuals and clinical populations~\cite{refSciRep2020,refPsycholMed2023}.

\subsection{Trial Structure and Temporal Constraints}

Each experimental trial follows a fixed and rapid sequence designed to impose strong temporal constraints on gaze estimation:

\begin{enumerate}
    \item a central fixation cross is displayed for a randomly jittered duration between 500 and 1500\,ms.
    \item the presentation of two images (left/right) occurs for 2 seconds,
    \item a brief validation phase is indicated by a small black dot, requiring a key press,
    \item returning to the fixation cross to initiate the next trial.
\end{enumerate}

This sequence is repeated in blocks, resulting in a total of 96 left–right image pairs per session. The task lasts approximately 10 minutes, with a short break halfway through. The short exposure duration of stimuli and the rapid alternation between fixation and free viewing make this task particularly challenging for webcam-based eye trackers, which must deliver stable predictions under tight timing constraints.

\begin{figure}[ht]
\centering
\subfigure[Fixation cross]{
    \includegraphics[width=0.30\linewidth]{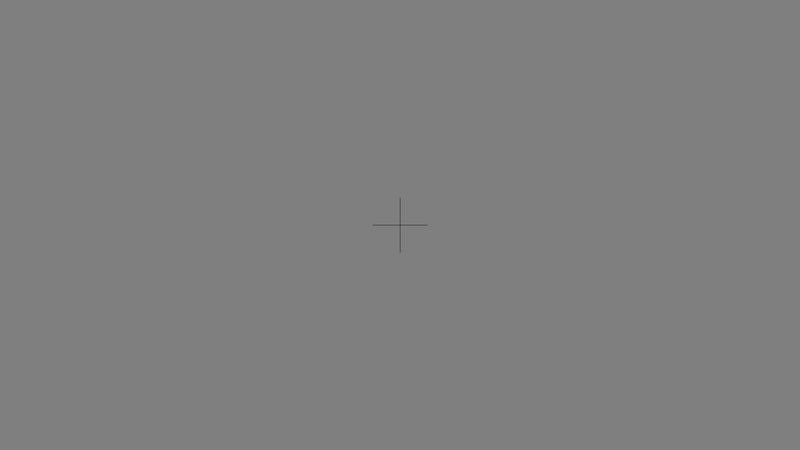}
}
\hfill
\subfigure[Stimulus pair presentation]{
    \includegraphics[width=0.30\linewidth]{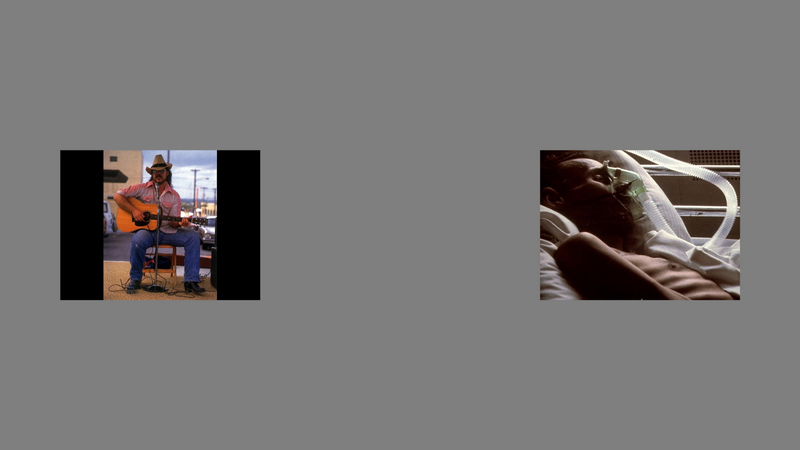}
}
\hfill
\subfigure[Target detection]{
    \includegraphics[width=0.30\linewidth]{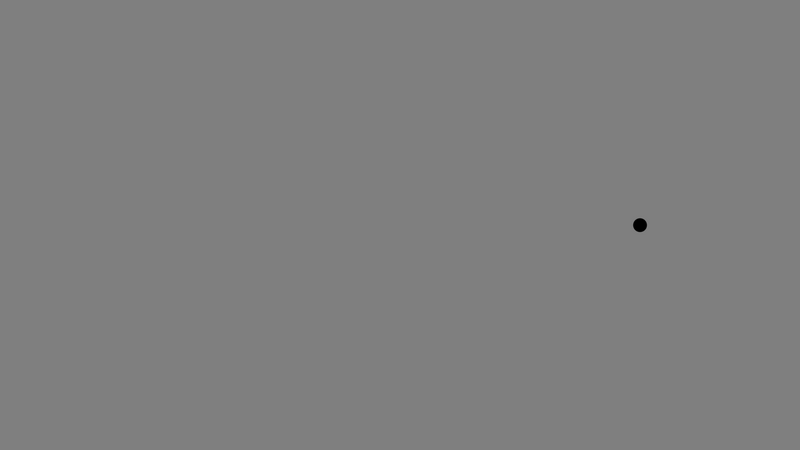}
}
\vspace{-2mm}
\caption{Example of a single trial in the Dot-Probe Task used for evaluation. Each trial starts with a central fixation cross, followed by the presentation of two stimuli (negative vs.\ neutral). One stimulus is then replaced by a target dot, which the participant must localize via a key press while gaze is continuously recorded.}
\label{fig:dpt_example}
\vspace{-2mm}
\end{figure}

\subsection{Participants}

Eighteen healthy participants took part in the study. All participants completed the full experimental protocol using the same experimental setup, namely a standard consumer laptop equipped with a built-in webcam, without a dedicated GPU or specialized hardware. Recordings were nevertheless conducted in different rooms and naturalistic conditions, resulting in variability in ambient lighting, participant posture, and the presence or absence of corrective glasses.

Eye movements were recorded throughout the task using two different webcam-based trackers: the commercial SeeSo SDK and EyeTheia, utilizing the pretrained iTracker model with user-specific calibration (Approach~1). Both trackers processed the same webcam input stream and were evaluated on identical trials, enabling a direct comparison under matched experimental conditions.

\subsection{Objective of the Comparison}

The purpose of this experiment is to assess whether EyeTheia can serve as a viable alternative to a commercial webcam-based eye tracker in a demanding experimental paradigm that has already been validated in experimental and clinical research. Rather than optimizing performance on synthetic benchmarks, this study evaluates tracking behavior in a realistic cognitive task involving rapid stimulus presentation, sustained attention, and prolonged recording sessions.

At the time of data collection, only the pretrained EyeTheia model (Approach~1) was deployed during the experimental sessions. The model trained from scratch on MPIIFaceGaze (Approach~2) was not included in this comparative evaluation due to the practical constraints of the experimental platform. In particular, the web-based implementation of the task required running multiple eye-tracking pipelines concurrently on the same consumer laptop, which imposed strict real-time and stability constraints. Under these conditions, integrating a third tracker did not allow for sufficiently stable execution during the experimental sessions.

Importantly, this exclusion does not reflect a limitation of the Approach~2 model itself, but rather the state of the experimental deployment at the time of data collection. The integration of the MPIIFaceGaze-based tracker into the experimental platform is part of ongoing work and is discussed in Section~\ref{sec:conclusion}.

The goal of this evaluation is not to provide a comprehensive benchmark against all gaze estimation systems, but to assess whether EyeTheia can serve as a practical alternative in a real experimental pipeline.
SeeSo is used as a reference as it is the production system employed in our protocol after internal validation for stability and usability.
Offline comparisons would not capture the real-time constraints, temporal dynamics, and system-level interactions central to our use case.

In the following section, we report quantitative comparisons between the three trackers, focusing on their ability to discriminate gaze allocation toward left versus right stimuli across trials.
\section{Results}

\subsection{Training and Calibration Results}
\label{sec:training_results}

We report here the main training and calibration results for the two EyeTheia approaches introduced in Section~\ref{sec:methodology}.
This section focuses on (i) the selection of training hyperparameters for the model trained from scratch on MPIIFaceGaze (Approach~2), (ii) the effect of user-specific calibration through fine-tuning for both approaches, and (iii) the validation performance of both models on MPIIFaceGaze.

\paragraph{Hyperparameter selection for Approach~2.}
For the model trained from scratch on MPIIFaceGaze, we investigated the impact of the Smooth~L1 (Huber) loss parameter~$\beta$ on validation performance.
A grid search was conducted over $\beta \in [0.01, 1.0]$, using a fixed learning rate and subject-disjoint validation splits.

\begin{figure}[t]
\centering
\subfigure[Validation loss curves for different values of $\beta$ across epochs.]{
    \includegraphics[width=0.47\linewidth]{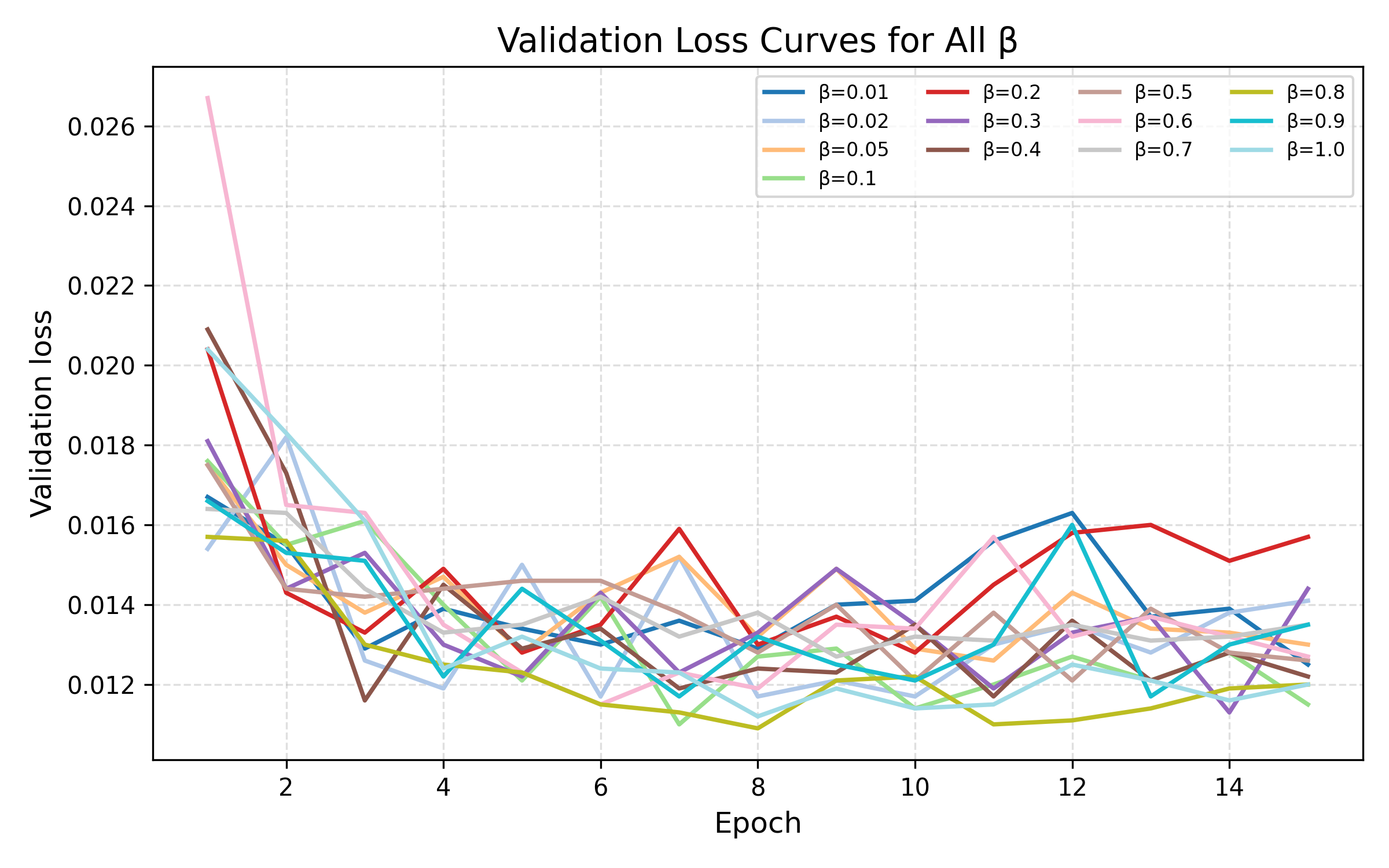}
    \label{fig:val_curves_all_beta}
}
\hfill
\subfigure[Best validation loss achieved for each $\beta$.]{
    \includegraphics[width=0.47\linewidth]{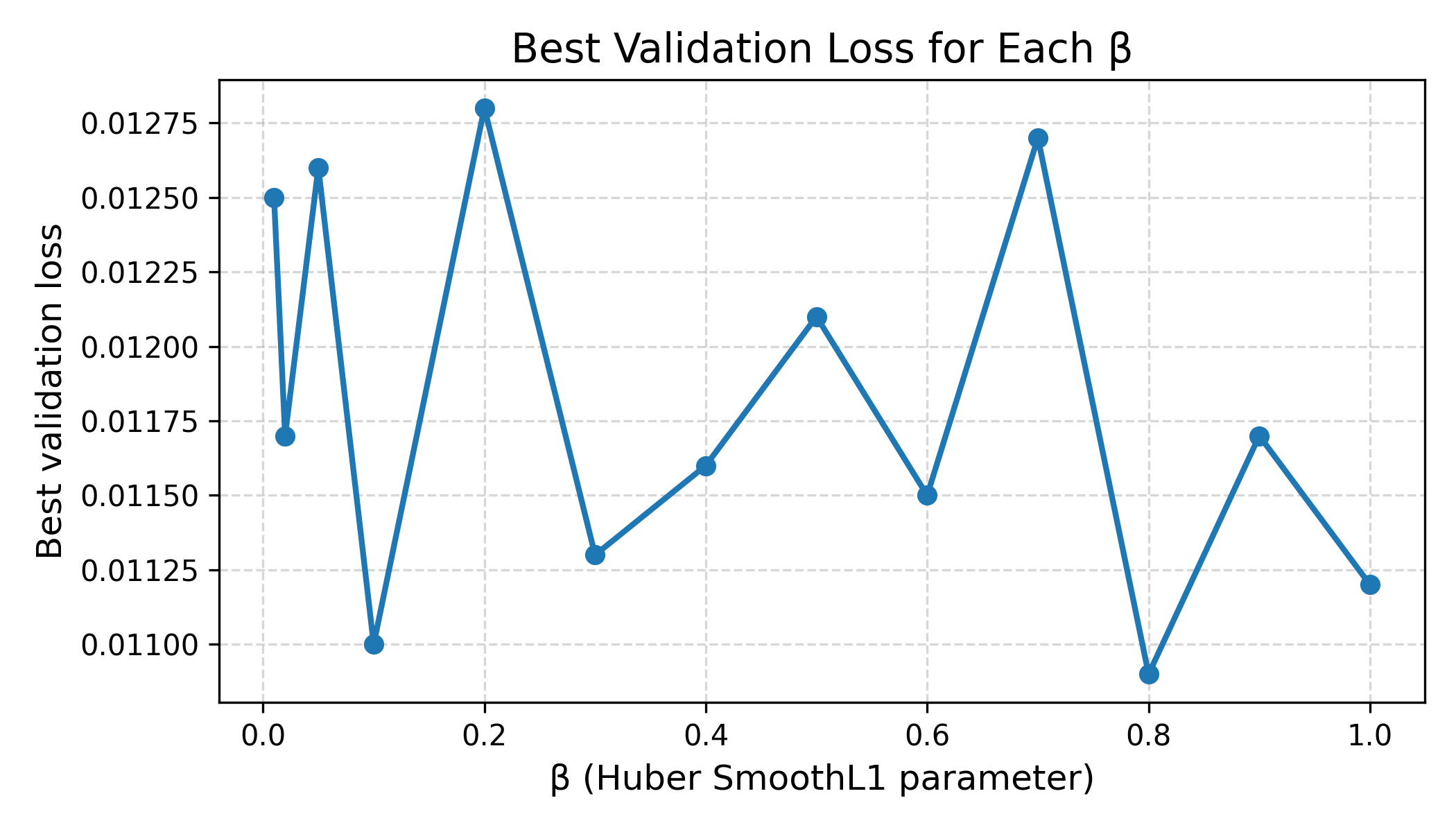}
    \label{fig:best_val_loss_beta}
}
\vspace{-2mm}
\caption{Effect of the Smooth~L1 (Huber) parameter $\beta$ on validation performance for Approach~2 trained on MPIIFaceGaze.
(\textbf{Left}) Validation loss trajectories over training epochs for all tested values of $\beta$.
(\textbf{Right}) Minimum validation loss achieved for each $\beta$, with the best performance obtained for $\beta = 0.8$.}
\label{fig:beta_selection}
\vspace{-2mm}
\end{figure}

Figure~\ref{fig:best_val_loss_beta} reports the best validation loss achieved for each $\beta$ value.
The lowest validation error is obtained for $\beta = 0.8$, which is therefore retained in all subsequent experiments.
This result highlights the importance of controlling the transition between the quadratic and linear regimes of the loss when regressing gaze coordinates, particularly in the presence of outliers.

\paragraph{Validation performance on MPIIFaceGaze.}
We additionally evaluate the predictive accuracy of both EyeTheia variants on held-out validation subjects from MPIIFaceGaze, using subject-disjoint splits.
This evaluation aims to quantify baseline gaze prediction performance prior to user-specific calibration and to compare the generalization behavior of the pretrained model (Approach~1) with the model trained from scratch (Approach~2).
The test set used for this evaluation corresponds to the same subject-disjoint split employed during the training of Approach~2, ensuring a consistent and comparable evaluation protocol across models.

Table~\ref{tab:mpiifacegaze_eval_px} reports quantitative results in pixel space on the MPIIFaceGaze test split.
Overall, both models achieve comparable performance across all reported metrics.
The pretrained model (Approach~1) yields slightly lower global errors in terms of RMSE and mean L2 distance; however, the differences between the two approaches remain modest.
These results indicate that, prior to user-specific calibration, both training strategies provide similar levels of accuracy on unseen subjects.

\begin{table}[t]
\centering
\caption{Validation performance on the MPIIFaceGaze test split (7\,910 samples).
Lower values indicate better performance ($\downarrow$).}
\label{tab:mpiifacegaze_eval_px}
\begin{tabular}{lcc}
\toprule
\textbf{Metric} & \textbf{Approach~1 (cm$\rightarrow$px)} & \textbf{Approach~2 (norm$\rightarrow$px)} \\
\midrule
RMSE$_{2D}$ (px)        & 428.15 & 447.37 \\
Mean L2 error (px)       & 392.89 & 395.53 \\
Mean L2 / screen diagonal (\%) & 23.92 & 24.05 \\
\bottomrule
\end{tabular}
\end{table}

\paragraph{Analysis of the performance gap.}
The small performance gap between Approach~1 and Approach~2 reflects a trade-off between dataset scale and domain alignment.
Approach~1 benefits from large-scale pretraining on GazeCapture, but suffers from a domain mismatch with laptop usage.
Approach~2 is better aligned with the target setting but limited by the smaller size of MPIIFaceGaze.
In addition, gaze estimation without user-specific calibration remains highly challenging due to strong inter-subject variability, which explains the similar error levels observed prior to calibration.

\paragraph{Effect of user-specific calibration.}
Both EyeTheia variants rely on a short, user-specific calibration phase during which the model is fine-tuned on a small set of calibration samples collected at the beginning of each session.
This calibration procedure is applied consistently to both the pretrained model (Approach~1) and the model trained from scratch (Approach~2).

\begin{figure}[t]
\centering

\begin{minipage}{0.48\linewidth}
    \centering
    \includegraphics[width=\linewidth]{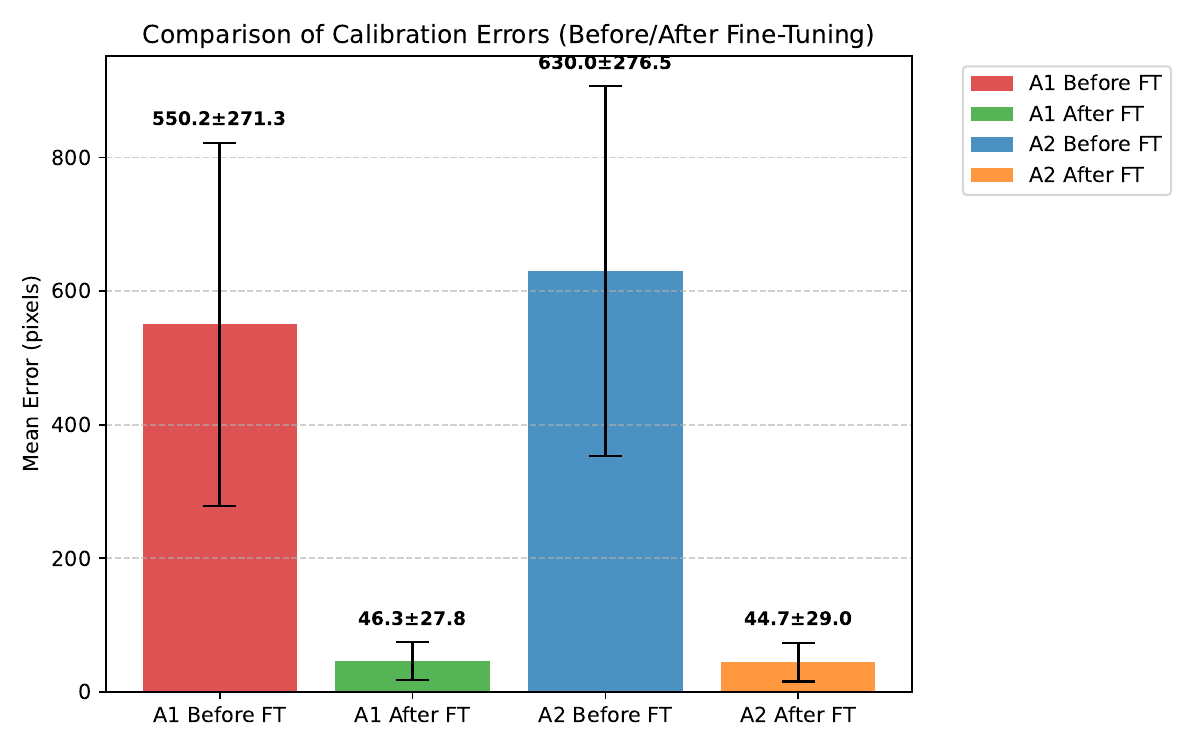}
    \caption{Comparison of gaze prediction errors before and after user-specific calibration.}
    \label{fig:calibration_comparison}
\end{minipage}
\hfill
\begin{minipage}{0.48\linewidth}
    \centering
    \includegraphics[width=\linewidth]{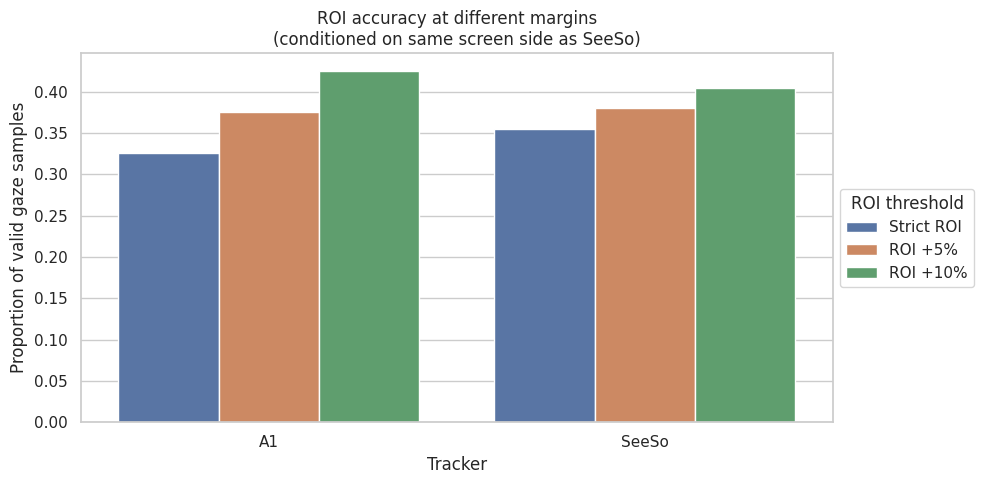}
    \caption{ROI-based accuracy for EyeTheia and SeeSo.}
    \label{fig:roi_accuracy}
\end{minipage}

\end{figure}

Figure~\ref{fig:calibration_comparison} summarizes the effect of this fine-tuning step by comparing gaze prediction errors before and after calibration.
For both approaches, calibration leads to a substantial reduction in pixel error, confirming the effectiveness of the proposed adaptation strategy.
Notably, while the pretrained model starts from a stronger initialization, the model trained from scratch also benefits significantly from calibration, thereby reducing the gap between the two approaches.

Overall, these results confirm that (i) careful loss tuning is necessary when training gaze estimation models from scratch, (ii) both approaches achieve meaningful validation performance on MPIIFaceGaze, and (iii) lightweight user-specific calibration constitutes a key component for achieving accurate gaze prediction under real-world conditions.

\subsection{Comparison with SeeSo}
\label{sec:results_seeso}

We compare EyeTheia (Approach~1) against the commercial SeeSo SDK using the Dot-Probe task described in Section~\ref{sec:experiments}, focusing on the ability to discriminate gaze allocation toward the left or right side of the screen during stimulus presentation.

\paragraph{Screen-side agreement.}
Screen-side agreement is computed by discretizing horizontal gaze positions relative to the screen center: samples with $x < W/2$ are assigned to the left side, and samples with $x \geq W/2$ to the right.
This metric is evaluated exclusively during stimulus presentation intervals.
Across all participants and trials, EyeTheia agrees with SeeSo on the predicted screen side during 75.1\% of valid samples.
This level of agreement indicates that EyeTheia reliably captures coarse left/right attentional shifts required by the Dot-Probe paradigm, despite increased sensitivity to short-lived fluctuations around stimulus onset.

\paragraph{ROI-based spatial consistency.}
We further evaluate spatial behavior using a region-of-interest (ROI) criterion centered on the displayed stimulus.
Figure~\ref{fig:roi_accuracy} reports accuracy under strict ROIs and relaxed ROIs expanded by 5\% and 10\%.
While SeeSo achieves slightly higher accuracy under the strict ROI (35.5\% vs.\ 32.6\%), performance converges as the ROI is relaxed, with EyeTheia reaching comparable or slightly higher accuracy under the 10\% margin (42.5\% vs.\ 40.4\%).
This trend suggests that EyeTheia reliably captures coarse left--right attentional allocation, with reduced fine-grained spatial precision attributable to short-term prediction noise.

\begin{figure}[t]
\centering
\includegraphics[width=0.7\linewidth]{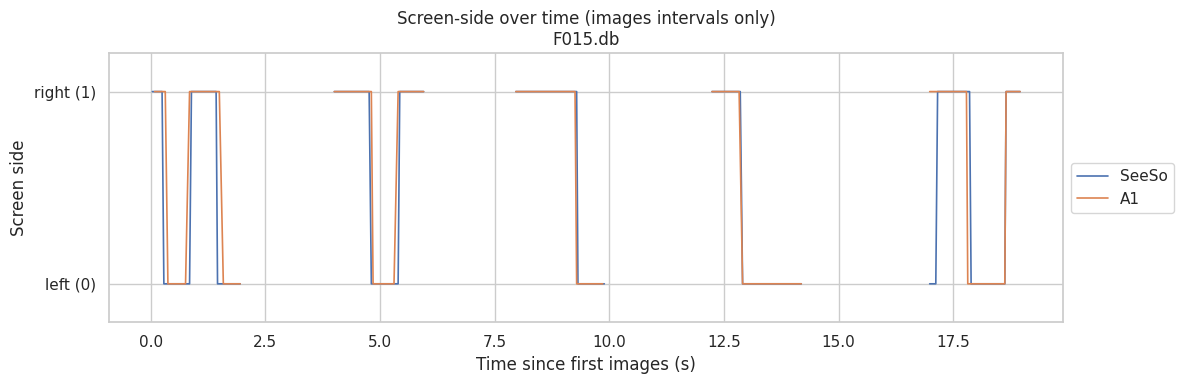}
\vspace{-2mm}
\caption{Temporal evolution of predicted screen side during stimulus presentation intervals for a representative participant.
Blue curves correspond to SeeSo predictions, while orange curves correspond to EyeTheia (Approach~1).
The plot illustrates the overall agreement in left/right gaze allocation, with occasional short-lived divergences near fixation--stimulus transitions.}
\label{fig:timeline_comparison}
\vspace{-2mm}
\end{figure}

\paragraph{Temporal behavior and stability.}
Temporal analysis shows that EyeTheia follows the same left--right alternation patterns as SeeSo across trials, with occasional short-lived divergences near transitions between fixation and stimulus presentation (Figure~\ref{fig:timeline_comparison}).
Quantitatively, EyeTheia exhibits higher inter-frame jitter than SeeSo (69.4 vs. 26.2 pixels), reflecting increased frame-to-frame variability in the absence of explicit temporal smoothing.

\begin{table}[t]
\centering
\caption{Quantitative comparison between SeeSo and EyeTheia (Approach~1) on the experimental task.}
\label{tab:seeso_comparison}
\begin{tabular}{lcc}
\toprule
Metric & EyeTheia (A1) & SeeSo \\
\midrule
Screen-side agreement (\%) & 75.1 & -- \\
Strict ROI accuracy & 0.326 & \textbf{0.355} \\
ROI +5\% accuracy & 0.375 & \textbf{0.381} \\
ROI +10\% accuracy & \textbf{0.425} & 0.404 \\
Mean jitter (pixels) & 69.4 & \textbf{26.2} \\
\bottomrule
\end{tabular}
\end{table}

\section{Conclusion}
\label{sec:conclusion}

In this work, we introduced \emph{EyeTheia}, an open and lightweight pipeline for webcam-based gaze estimation designed for browser-based experimental platforms and real-world cognitive and clinical research.
Built upon the iTracker architecture, EyeTheia combines landmark-driven feature extraction using MediaPipe, deep learning–based gaze regression, and a short user-specific calibration phase, enabling real-time deployment on standard consumer hardware.

We investigated two complementary strategies: adapting a pretrained iTracker model originally trained on mobile data (Approach~1) and training the same architecture from scratch on MPIIFaceGaze to obtain a desktop-oriented baseline (Approach~2).
Validation results show that both approaches achieve comparable performance on unseen subjects prior to calibration, while user-specific fine-tuning consistently yields substantial error reductions, highlighting the importance of lightweight adaptation in unconstrained settings.

Beyond dataset-based evaluation, EyeTheia was assessed in a demanding experimental task based on the Dot-Probe paradigm.
In comparison with a commercial webcam-based tracker (SeeSo), EyeTheia reliably discriminates between left and right gaze during stimulus presentation, capturing the coarse attentional signals required by established cognitive paradigms.
Although higher temporal jitter and reduced fine-grained spatial precision are observed, these differences primarily affect short-term stability rather than global gaze allocation.

The current pipeline does not yet incorporate explicit temporal smoothing, contributing to increased frame-to-frame variability, and only the pretrained model was deployed during experimental sessions due to the real-time constraints of the web-based platform.
These limitations reflect practical deployment conditions rather than intrinsic model deficiencies.

Future work will focus on improving temporal stability through explicit filtering in the inference stream, such as the \emph{One Euro Filter}~\cite{OneEuroFilter}, and on extending the architecture to better exploit geometric cues from facial landmarks, including depth-related and pose-dependent information.
Overall, EyeTheia provides a transparent and extensible foundation for low-cost gaze tracking, bridging the gap between laboratory-grade systems and scalable, reproducible behavioral assessment on consumer devices.

%
%
\bibliographystyle{splncs04}

\bibliography{refs}
\end{document}